\documentclass[conference, 10pt, letterpaper]{ieeeconf}
\IEEEoverridecommandlockouts
\usepackage{amsmath,amssymb,amsfonts}
\usepackage{xcolor}
\usepackage{todonotes}
\usepackage{multirow}
\usepackage{booktabs}
\usepackage{censor}

\usepackage{nick_macros}
\newcommand{\reals}[1]{\mathbb{R}^{#1}}

\makeatletter
\let\NAT@parse\undefined
\makeatother
\usepackage[hidelinks]{hyperref}
\usepackage[capitalize]{cleveref}
\usepackage{cite}

\begin{document}

\title{\LARGE \bf
Tendon-Actuated Robots with a Tapered, Flexible Polymer Backbone: Design, Fabrication, and Modeling*
}

\author{Harald Minde Hansen$^{1}$, Nandita Gallacher$^{1}$, Nicholas B. Andrews$^{2}$,\\ Kristin Y. Pettersen$^{3}$, Jan Tommy Gravdahl$^{3}$, and Mario di Castro$^{1}$
\thanks{*This work was supported in part by {ONR Award N00014-23-1-2171}}%
\thanks{$^{1}${Mechatronics, Robotics and Operation Section, European Organization for Nuclear Research (CERN), Meyrin, Switzerland {\tt\small [{\href{mailto:harald.minde.hansen@cern.ch}{harald.minde.hansen}, \href{mailto:nandita.gallacher@cern.ch}{nandita.gallacher}, \href{mailto:mario.di.castro@cern.ch}{mario.di.castro}}]@cern.ch}}}%
\thanks{$^{2}${Department of Aeronautics and Astronautics, University of Washington, Seattle, WA, USA {\tt\small {\href{mailto:nian6018@uw.edu}{nian6018@uw.edu}}}}}%
\thanks{$^{3}${Department of Engineering Cybernetics, Norwegian University of Science and Technology (NTNU), Trondheim, Norway
{\tt\small [{\href{mailto:kristin.y.pettersen@ntnu.no}{kristin.y.pettersen}, \href{mailto:jan.tommy.gravdahl@ntnu.no}{jan.tommy.gravdahl}}]@ntnu.no}}}
}

\maketitle

\begin{abstract}
This paper presents the design, modeling, and fabrication of 3D-printed, tendon-actuated continuum robots with a flexible, tapered backbone made from thermoplastic polyurethane (TPU). Unlike many continuum robots that are single-purpose and costly, the proposed design prioritizes customizability, rapid assembly, and low cost, while geometric tapering enables high curvature and enhanced distal compliance, supporting a broad range of compliant inspection and manipulation tasks. The scalable design includes an integrated electronics base housing that provides direct tendon tension control and sensing through actuators and compression load cells. The tapered backbone is modeled using a generalized forward kinetostatic model based on Cosserat rod theory that explicitly accounts for spatially varying cross-sectional geometry, enabling systematic exploration of the configuration space as a function of the geometric design parameters. To capture nonlinear material properties and manufacturing uncertainties, measured cable tensions serve as a proxy variable in a scheduling method that allows the Young's modulus to vary with tension. The model is validated against motion capture data, achieving centimeter-level shape prediction accuracy after calibration. Finally, an accompanying video demonstrates teleoperated grasping with an endoscopic gripper routed along the continuum robot while mounted on a 6-DoF robotic arm, and parameterized iLogic/CAD scripts are provided for rapid geometry generation and scaling.
\end{abstract}

\section{INTRODUCTION}\label{sec:introduction}



The inherent compliance of continuum robots enables intrinsically safe interaction between the robot and the environment. This property is crucial when operating in confined spaces cluttered with electrical cables and sensitive equipment. The tendon actuation strategy together with a flexible backbone transmission unit benefits from compactness, simple fabrication, well-studied robot models, and high force transmission. A common approach for tendon actuation is to use a flexible backbone connected to either rigid discs \cite{qi2019, till2019a, shen2020} or a supporting structure \cite{moralesbieze2020}, in which the tendons are threaded through holes. A sufficiently stiff backbone prevents the structure from collapsing under zero tendon tension \cite{kim2014}. Unlike pneumatic and conductivity-driven actuation, this approach eliminates leakage risks \cite{walker2020} and operates without high voltages \cite{jacquemin2023}.

\begin{figure}
\centerline{\includegraphics[width=\linewidth]{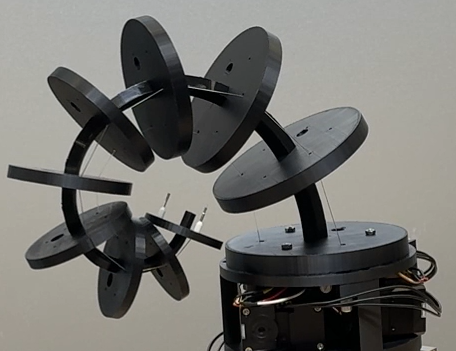}}
\caption{The proposed 3D-printed continuum robot design featuring a tapered, flexible TPU backbone, an integrated electronics base housing, and three tendons equally spaced along the circumference.}
\label{fig:flex_backbone}
\vspace{-0.5cm}
\end{figure}

Backbone tapering enables greater curvature under tip loading than a uniform-diameter backbone. Combined with tendon actuation, it yields a system whose stiffness decreases toward the tip, providing a stable, stiff base while allowing greater dexterity and compliance distally. For a tendon-actuated continuum robot, this expanded configuration space enables the robot to envelop objects more tightly near its tip, facilitating the grasping of small objects. The ability to perform multiple rotations into a spiral configuration further increases inward contact forces near the tip, improving grasp security. These benefits are enhanced further when tapering is applied to a backbone manufactured from a compliant material.

TPU has become an increasingly attractive material for 3D-printed engineering designs owing to its affordability, modularity, and ease of fabrication, enabling novel applications such as structural mesh design \cite{Masmeijer2026-qu}. A relevant example is presented in \cite{wang2025}, which explores a biologically inspired design in which compliant, tapered robots are fabricated from TPU and actuated via tendons. Through geometric locking of backbone sections, their design reliably produces a logarithmic spiral when fully curled, enabling impressive grasping capabilities in both standalone and multi-agent configurations. 
However, a model-based, curvature-driven design optimization for this class of robots has not yet been explored. 


Cosserat rod theory, combined with tendon actuation modeled through either a Newtonian \cite{rucker2011, till2019a} or Lagrangian \cite{tummers2023, boyer2021} approach, provides a basis for deriving forward dynamic and kinetostatic models of continuum robots with flexible backbones. These formulations, however, do not explicitly account for variations in cross-sectional area along the rod length, though spatially varying stiffness profiles for Cosserat rods have been addressed for multi-material soft robots in \cite{esser2025}. While higher-fidelity finite element methods can model spatially varying geometries more accurately \cite{weeger2018}, such approaches typically yield complex models that do not admit compact ordinary differential equation formulations and incur significant computational cost.

Experimental validation of models has been performed predominantly on robots with thin metal backbones, such as steel spring \cite{till2019a, rone2014} or nickel-titanium alloy \cite{wang2025, kheradmand2025}, which are typically designed for single-purpose, high-precision applications such as medical interventions \cite{dupont2022, burgner-kahrs2015} and exhibit behavior that differs substantially from compliant polymer structures such as TPU. Accurately modeling TPU presents considerable challenges: its stress-strain behavior is nonlinear even at small strains \cite{Masmeijer2026-qu, marco2025}, and micro-structural defects and anisotropy from the FDM printing process cause the effective Young's modulus to vary spatially.




The primary contribution of this work is a novel tendon-actuated continuum robot design with a tapered TPU backbone. Compared to existing designs, the proposed system is distinguished by its simplicity, rapid assembly, accessibility, and adaptability to a wide range of applications at minimal cost. An integrated electronics housing at the base of the backbone contains the tendon actuation system, motor encoders for measuring cable displacement, and load cells for measuring tendon tension. The modular design allows the robot to be mounted as an end-effector on a rigid manipulator or integrated with endoscopic tools, serving as both a compliant tool carrier and an inspection instrument.

The robot is modeled using the approach in \cite{esser2025}, and the model serves as a practical design tool that enables practitioners to select geometric parameters yielding a desired configuration space. This capability is demonstrated through a taper angle sweep and a backward mapping from desired curvature profiles to optimal backbone taper angles. Model validity is assessed experimentally using a camera-based motion capture system together with load cell measurements of tendon tension. To account for material nonlinearities, manufacturing variability, and other unmodeled effects, the model is calibrated against motion capture data from the physical robot, with the Young's modulus varied through a scheduling approach conditioned on the measured tendon tensions to minimize shape prediction error. Finally, a parametric CAD synthesis tool that automatically maps user-specified geometric parameters to fully instantiated, properly scaled robot models is provided.


The remainder of the paper is organized as follows: \cref{sec:design} outlines our design and fabrication contributions to flexible backbone robots, the actuation system, and the endoscopic tool integration. \cref{sec:model} develops the Cosserat rod model of the tapered backbone with a design sensitivity analysis. \cref{sec:model_validation} assesses the model through simulations and experiments, and \cref{sec:conclusion} presents conclusions and future work. Finally, the \hyperref[appendix]{Appendix} presents open-loop teleoperated scenarios with endoscopic tool and camera integration in an accompanying video, demonstrating the adaptability of the robot design. The \hyperref[appendix]{Appendix} also provides modeling code and Inventor iLogic scripts that automate CAD generation from high-level parameters such as backbone length, base width, taper angle, and disc count.

\section{DESIGN AND FABRICATION}\label{sec:design}
This section presents our design of a scalable, 3D-printed soft robot with a tapered, flexible polymer backbone and rigid discs, shown in \cref{fig:flex_backbone}. The automated design process for the 3D-printed components is discussed first, followed by the integrated electronics base housing, which provides tendon tension actuation and sensing through actuators and compression load cells.

\subsection{Automated Backbone Design}
Motivated by the biologically inspired designs explored in \cite{wang2025}, the backbone is tapered such that it forms a logarithmic spiral when fully curled, enabling applications in body grasping and confined-space inspection. In our design, the disc radii, thicknesses, and spacing vary according to the same logarithmic ratio. The discs are friction-fit to the backbone, and each disc features three equidistant tendon-routing holes arranged about the backbone circumference, as well as an additional, unused hole of fixed radius to accommodate an endoscopic tool and camera.

Rather than relying on a static CAD model of the robot, the fabrication process is automated in the rules-based design software Inventor iLogic to allow for rapid prototyping and iterative design modifications. Our implementation automatically generates a CAD file based on the backbone length, taper angle, number of discs, base diameter, cable hole diameter, and other physical parameters that determine the stiffness, structural strength, and configuration space of the robot. A user can update these values in an Excel file tailored to the intended robot performance and operating environment, then run the iLogic script to rapidly generate a part file ready for 3D printing.

\subsection{Electronics Housing and Integration}
\begin{figure}
    \vspace{0.15cm}
    \centerline{\includegraphics[scale=.5]{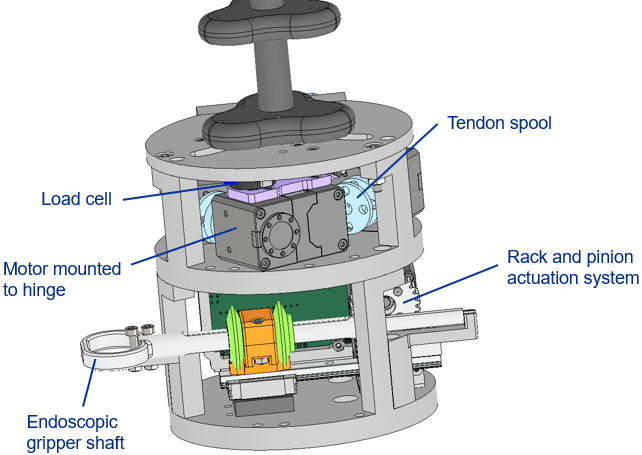}}
    \caption{Actuation and sensing electronics housing located at base of the tapered backbone with a subspace for endoscopic tool actuation.}
    \label{fig:actuation_system}
\end{figure}

The actuation and sensing system is housed inside a hollow cylindrical enclosure at the base of the backbone, shown in \cref{fig:actuation_system}. The base housing is designed to operate independently or to be mounted on the end-effector flange of a rigid manipulator or mobile robot, depending on the intended application. Three Dynamixel XH430-210T motors are mounted on hinges and spaced equidistantly about the inner circumference of the housing. Each motor drives a spool around which a tendon coils; when in tension, the tendon pulls the hinged motor perpendicular to the base of the frame, such that the hinge pushes the compression point of an FX29 load cell that is mounted to the actuation system frame, as shown in \cref{fig:actuation_system}. This facilitates accurate shape estimation and allows torque control to be mapped directly to tendon tension control. The tendons are routed through their corresponding disc holes and fastened at the distal disc. 

\section{KINETOSTATIC MODEL}\label{sec:model}
This section introduces the fundamentals of Cosserat rod theory and its application to tendon-actuated continuum robots with tapered, flexible backbones. To capture the effects of tapering, the proposed model allows the cross-sectional area and second moments of area to vary continuously along the rod's curvilinear arc length parameter \cite{esser2025}.

\begin{figure}
    \vspace{0.15cm}
    \centerline{\includegraphics[scale=.45]{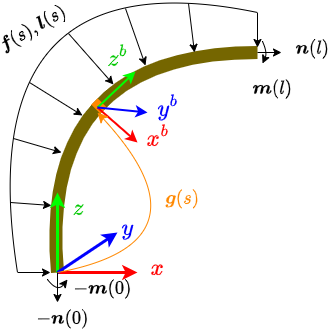}}
    \caption{Cosserat rod free body diagram where $\{x, y, z\}$ denotes the coordinate frame attached to the proximal (mounted) end of the robot, and $\{x^b, y^b, z^b\}$ denotes the local coordinate frame at backbone arc length $s$. The action $g(s)$ represents the transformation from the proximal end to the cross-section pose at $s$.}
    \label{fig:cosserat_rod}
    \vspace{-0.5cm}
\end{figure}

\subsection{Cosserat Rod Model}
Cosserat rods are represented as a continuous stack of infinitesimal rigid cross-sections, parameterized by arc length $s \in [0, l]$, where $l$ is the total length of the rod. A free body diagram for a Cosserat rod is shown in \cref{fig:cosserat_rod}. The centerline curve of the rod with respect to the inertial frame is parameterized by the homogeneous transformation matrix
\begin{equation}
\label{RodTransformation}
    \boldsymbol{g}(s) = \begin{bmatrix}
        \boldsymbol{R}(s) & \boldsymbol{p}(s)\\
        0 & 1
    \end{bmatrix} \in \text{SE}(3),
\end{equation}
where $\boldsymbol{R}(s) \in \text{SO}(3)$ and $\boldsymbol{p}(s) \in \reals{3}$ denote the orientation and position of the centerline of the rod in $s$. The evolution of $\boldsymbol{g}(s)$ along the rod is described by the linear ($\boldsymbol{v}(s) \in \reals{3}$) and angular ($\boldsymbol{u}(s) \in \reals{3}$) rate of change vectors such that
\begin{subequations}
    \label{RodKinematics}
    \begin{align}
        \dot{\boldsymbol{p}}(s) &= \boldsymbol{R}(s)\boldsymbol{v}(s), \\
        \dot{\boldsymbol{R}}(s) &= \boldsymbol{R}(s)\widehat{\boldsymbol{u}}(s).
    \end{align}
\end{subequations}
Following the notation of \cite{rucker2011}, the overdot notation ($\dot{x}$) denotes the derivative with respect to the curvilinear abscissa $s$. The hat operator $\widehat{\boldsymbol{a}}$ and its inverse $\boldsymbol{a}^{\vee}$ for a vector $\boldsymbol{a} = \begin{bmatrix} a_x & a_y & a_z \end{bmatrix}^\top$ are defined as
\begin{equation}
\label{SkewSymmetric}
    \widehat{\boldsymbol{a}} = \begin{bmatrix}
        0 & -a_z & a_y\\
        a_z & 0 & -a_x\\
        -a_y & a_x & 0
    \end{bmatrix}, \quad \left( \widehat{\boldsymbol{a}} \right)^{\vee} = \boldsymbol{a}.
\end{equation}


The Cosserat rod is in static equilibrium under distributed external force and couple density fields per unit $s$, $\boldsymbol{f}(s) \in \reals{3}$ and $\boldsymbol{l}(s) \in \reals{3}$, balanced by internal forces and couples, \mbox{$\boldsymbol{n}(s) \in \reals{3}$} and $\boldsymbol{m}(s) \in \reals{3}$, all expressed in the inertial frame. The static equilibrium equations are
\begin{subequations}
\label{EquilibriumEquations}
    \begin{align}
        &\dot{\boldsymbol{n}}(s) + \boldsymbol{f}(s) = \boldsymbol{0}, \\
        &\dot{\boldsymbol{m}}(s) + \dot{\boldsymbol{p}}(s)\times\boldsymbol{n}(s) + \boldsymbol{l}(s) = \boldsymbol{0}.
    \end{align}
\end{subequations}
In our model, we define the undeformed nominal configuration as straight and cylindrically tapered, such that \mbox{$\boldsymbol{v}_0 = \begin{bmatrix}
    0 & 0 & 1
\end{bmatrix}^\top$}, \mbox{$\boldsymbol{u}_0 = \begin{bmatrix}
    0 & 0 & 0
\end{bmatrix}^\top$}, and the undeformed transformation $\boldsymbol{g}_0(s) \in \text{SE}(3)$ is defined with the $+z$ axis in the distal direction and $+x$ pointing radially towards one of the tendons, which will be introduced in the next subsection.

Assuming linear elasticity, the constitutive laws relate internal forces and couples to the linear and angular strain fields, providing a first-order approximation suitable for moderate curvature and quasi-static loading:
\begin{subequations}
\label{ConstitutiveLaws}
    \begin{align}
        \boldsymbol{n}(s) &= \boldsymbol{R}(s)\boldsymbol{K}_{se}(s)(\boldsymbol{v}(s) - \boldsymbol{v}_0), \\
        \boldsymbol{m}(s) &= \boldsymbol{R}(s)\boldsymbol{K}_{bt}(s)(\boldsymbol{u}(s) - \boldsymbol{u}_0).
    \end{align}
\end{subequations}
The matrices $\boldsymbol{K}_{se}(s) \in \reals{3 \times 3}$ and $\boldsymbol{K}_{bt}(s) \in \reals{3 \times 3}$ are spatially varying stiffness matrices for shear and extension, and bending and torsion, respectively, and are defined as
\begin{subequations}
\label{StiffnessMatrices}
    \begin{align}
        \boldsymbol{K}_{se}(s) &= \text{diag}(GA(s), GA(s), EA(s)), \\
        \boldsymbol{K}_{bt}(s) &= \text{diag}(EI_{xx}(s),EI_{yy}(s), EI_{zz}(s)),
    \end{align}
\end{subequations}
where the function $\text{diag}(\cdot)$ places the inputs along the diagonal and off-diagonal terms are zero. Their spatial derivatives are given by
\begin{subequations}
\label{StiffnessMatricesDot}
    \begin{align}
        \dot{\boldsymbol{K}}_{se} &= \text{diag}(G\dot{A}(s), G\dot{A}(s), E\dot{A}(s)), \\
        \dot{\boldsymbol{K}}_{bt} &= \text{diag}(E\dot{I}_{xx}(s), E\dot{I}_{yy}(s), E\dot{I}_{zz}(s)).
    \end{align}
\end{subequations}
Here, $A(s)$ is the cross-sectional area, $E$ is Young's modulus, $G$ is the shear modulus, and $I_{xx}$, $I_{yy}$ and $I_{zz}$ are the second moments of area of about the axes $x$, $y$, and $z$, respectively.



Now, substituting the constitutive laws \eqref{ConstitutiveLaws} and the kinematic parametrization \eqref{RodKinematics} into the equilibrium equations \eqref{EquilibriumEquations} yields the following system of differential equations, which we refer to as the generalized model equations in forward kinetostatic form:
\begin{subequations}
\label{ExplicitModelEquations}
    \begin{align}
        \dot{\boldsymbol{p}} &= \boldsymbol{R}\boldsymbol{v}, \\
        \dot{\boldsymbol{R}} &= \boldsymbol{R}\widehat{\boldsymbol{u}}, \\
        \dot{\boldsymbol{v}} &= \dot{\boldsymbol{v}}_0 - \boldsymbol{K}_{se}^{-1}\left[ (\widehat{\boldsymbol{u}}\boldsymbol{K}_{se} + \dot{\boldsymbol{K}}_{se})(\boldsymbol{v} - \boldsymbol{v}_0) + \boldsymbol{R}^T\boldsymbol{f} \right], \\
        \dot{\boldsymbol{u}} &= \dot{\boldsymbol{u}}_0 - \boldsymbol{K}_{bt}^{-1} [(\widehat{\boldsymbol{u}}\boldsymbol{K}_{bt} + \dot{\boldsymbol{K}}_{bt})(\boldsymbol{u} - \boldsymbol{u}_0) \nonumber\\
        & \quad + \widehat{\boldsymbol{v}}\boldsymbol{K}_{se}(\boldsymbol{v} - \boldsymbol{v}_0) + \boldsymbol{R}^T\boldsymbol{l} ].
    \end{align}
\end{subequations}
Unlike constant-stiffness formulations, the spatial variation of  $\boldsymbol{K}_{se}(s)$ and $\boldsymbol{K}_{bt}(s)$ introduces additional $\dot{\boldsymbol{K}}_{se}(s)$ and $\dot{\boldsymbol{K}}_{bt}(s)$ terms in \eqref{ExplicitModelEquations}, which are essential for consistent modeling of tapered geometries. The dependence on $s$ is implied by the preceding derivations and omitted from \eqref{ExplicitModelEquations} for brevity. The graded stiffness profile induced by tapering mitigates localized curvature concentrations near the base, thereby improving numerical stability when solving the resulting boundary value problem. 

The rod is assumed to be clamped at its base $s=0$ and subjected to an external tip force $\boldsymbol{f}_l \in \reals{3}$ and couple $\boldsymbol{l}_l \in \reals{3}$ at $s=l$. These conditions yield the boundary conditions $\boldsymbol{p}(0) = 0$, $\boldsymbol{R}(0) = I_3$, $\boldsymbol{n}(l) = \boldsymbol{f}_l$, and $\boldsymbol{m}(l) = \boldsymbol{l}_l$, which together fully define the boundary value problem for the Cosserat rod of non-uniform cross-section.

\subsection{Tendon Actuation}\label{sec:model_augmentation}

With the boundary value problem fully defined for the tapered backbone, the kinetostatic model can be augmented to include the effects caused by tendon forces and couples to obtain a complete tendon-actuated robot model. The derivation follows a similar process to \cite{rucker2011} and the Newtonian approach outlined in \cite{tummers2023}, but differs as a consequence of the spatially varying stiffness matrices. Similar to the convergent tendon routing scenario in \cite{tummers2023}, the tendons are assumed to follow arbitrary non-straight paths, making the model representative of a broad class of robot backbone designs. Additionally, the tendons are assumed to be inextensible, frictionless, and follow continuous tendon routing paths rather than piecewise continuity between the discs.


The distributed loads on the Cosserat rod are decomposed into
\begin{subequations}
\label{ForceAndCoupleDecomposition}
    \begin{align}
        \boldsymbol{f}(s) &= \boldsymbol{f}_{act}(s) + \boldsymbol{f}_{ext}(s), \\
        \boldsymbol{l}(s) &= \boldsymbol{l}_{act}(s) + \boldsymbol{l}_{ext}(s),
    \end{align}
\end{subequations}
where $\boldsymbol{f}_{act}(s) \in \reals{3}$ and $\boldsymbol{l}_{act}(s) \in \reals{3}$ denote the forces and couples caused by the tendon actuation, while $\boldsymbol{f}_{ext}(s) \in \reals{3}$ and $\boldsymbol{l}_{ext}(s) \in \reals{3}$ are external forces and couples from the environment. From the action-reaction principle, the sums of forces and couples transmitted by each tendon $i$ to the backbone $\boldsymbol{f}_i(s) \in \reals{3}$ and $\boldsymbol{l}_i(s) \in \reals{3}$ are given by
\begin{subequations}
\label{TendonForcesAndCouples}
    \begin{align}
        \boldsymbol{f}_{act}(s) &= -\sum_{i=1}^m \boldsymbol{f}_i(s), \\
        \boldsymbol{l}_{act}(s) &= -\sum_{i=1}^m   \boldsymbol{R}(s)\boldsymbol{d}_i(s) \times \boldsymbol{f}_i(s),
    \end{align}
\end{subequations}
where $m$ is the total number of tendons, and \mbox{$\boldsymbol{d}_i(s) = \begin{bmatrix}
    d_{i,x}(s) & d_{i,y}(s) & 0
\end{bmatrix}^\top$} is the position of tendon $i$ in the cross-sectional frame $b$; as illustrated in \cref{fig:cosserat_rod}. 

We assume that the tendons are ideal strings that are perfectly flexible and obey constant tension with respect to $s$. Substituting the tendon tensions into the equilibrium equations \eqref{EquilibriumEquations} for each tendon yields
\begin{equation}
\label{TendonEquilibriumEquation}
    \boldsymbol{f}_i(s) = - \dot{\boldsymbol{n}}_i(s) = \tau_i \dot{\boldsymbol{t}}_i(s),
\end{equation}
where $\boldsymbol{n}_i(s) \in \reals{3}$ is the internal force at static equilibrium, $\tau_i \in \reals{}$ is the constant tension, and $\boldsymbol{t}_i(s) \in \reals{3}$ is the directional tangent, all with respect to tendon $i$. The centerline position of tendon $i$ in the inertial frame, $\boldsymbol{p}_i(s) \in \reals{3}$, is
\begin{equation}
\label{TendonCenterline}
    \boldsymbol{p}_i(s) = \boldsymbol{p}(s) + \boldsymbol{R}(s)\boldsymbol{d}_i(s),
\end{equation}
where we recall $\boldsymbol{p}(s)$ is the centerline position of the Cosserat rod modeled flexible backbone. For augmenting the unactuated Cosserat rod model to a tendon-actuated model, we will need the two derivatives of the tendon positions:
\begin{subequations}
    \begin{align}
        \dot{\boldsymbol{p}}_i(s) &= \boldsymbol{R}(s)(\widehat{\boldsymbol{u}}(s)\boldsymbol{d}_i(s) + \dot{\boldsymbol{d}}_i(s) + \boldsymbol{v}(s)), \\
        \ddot{\boldsymbol{p}}_i(s) &= \boldsymbol{R}(s)(\widehat{\boldsymbol{u}}(s)(\widehat{\boldsymbol{u}}(s)\boldsymbol{d}_i(s) + \dot{\boldsymbol{d}}_i(s) + \boldsymbol{v}(s)) \nonumber\\
        & \quad + \widehat{\dot{\boldsymbol{u}}}(s)\boldsymbol{d}_i(s) + \widehat{\boldsymbol{u}}(s)\dot{\boldsymbol{d}}_i(s) + \ddot{\boldsymbol{d}}_i(s) + \dot{\boldsymbol{v}}(s)).
    \end{align}
\end{subequations}
The normalized tendon tangent $\boldsymbol{t}_i(s) \in \reals{3}$ is defined as
\begin{equation}
\label{TendonTangent}
    \boldsymbol{t}_i(s) = \frac{\dot{\boldsymbol{p}}_i(s)}{||\dot{\boldsymbol{p}}_i(s)||}.
\end{equation}
The force distribution of tendon $i$ is defined as
\begin{equation}
\label{TendonDistributedForce}
    \boldsymbol{f}_i(s) = -\dot{\boldsymbol{n}}_i = \tau_i \frac{\widehat{\dot{\boldsymbol{p}}}_i^2}{||\dot{\boldsymbol{p}}_i||^3}\Ddot{\boldsymbol{p}}_i,
\end{equation}
and is derived in Appendix B of \cite{rucker2011} from \eqref{TendonCenterline} and \eqref{TendonTangent}.
Lastly, substituting \eqref{TendonDistributedForce} into \eqref{TendonForcesAndCouples} yields
\begin{subequations}
    \begin{align}
        \boldsymbol{f}_{act}(s) &= -\sum_{i=1}^m \tau_i \frac{\widehat{\dot{\boldsymbol{p}}}_i^2}{||\dot{\boldsymbol{p}}_i||^3}\ddot{\boldsymbol{p}}_i, \\
        \boldsymbol{l}_{act}(s) &= -\sum_{i=1}^m \boldsymbol{R}(s)\boldsymbol{d}_i(s) \times \tau_i \frac{\widehat{\dot{\boldsymbol{p}}}_i^2}{||\dot{\boldsymbol{p}}_i||^3}\ddot{\boldsymbol{p}}_i.
    \end{align}
\end{subequations}
Using the previously derived Cosserat rod model, the effects of tendon actuation can now be incorporated, yielding a forward kinetostatic model that predicts the continuum robot configuration from a given set of tendon forces. Once again, the dependence on $s$ is omitted for brevity in the following series of equations. 

First, we define
\begin{align}
    \dot{\boldsymbol{p}}_i = \boldsymbol{R}\dot{\boldsymbol{p}}_i^b, \quad
    \ddot{\boldsymbol{p}}_i = \boldsymbol{R}\ddot{\boldsymbol{p}}_i^b,
\end{align}
where $\dot{\boldsymbol{p}}_i^b$ and $\Ddot{\boldsymbol{p}}_i^b$ are the centerline position derivatives of tendon $i$ with respect to $s$, expressed in the cross-sectional frame $b$. Now, we define the following auxiliary variables \cite{rucker2011}:
\begin{subequations}
\label{Identities}
    \begin{align}
        \boldsymbol{A}_i &= -\tau_i\frac{(\widehat{\dot{\boldsymbol{p}}}_i^b)^2}{||\dot{\boldsymbol{p}}_i^b||^3}, \quad \boldsymbol{A} = \sum_{i=1}^m \boldsymbol{A}_i, \\
        \boldsymbol{B}_i &= \widehat{\boldsymbol{d}}_i\boldsymbol{A}_i, \quad \boldsymbol{B} = \sum_{i=1}^m \boldsymbol{B}_i, \\
        \boldsymbol{G} &= -\sum_{i=1}^m \boldsymbol{A}_i\widehat{\boldsymbol{d}}_i, \quad \boldsymbol{H} = -\sum_{i=1}^m \boldsymbol{B}_i\widehat{\boldsymbol{d}}_i, \\
        \boldsymbol{a}_i &= \boldsymbol{A}_i(\widehat{\boldsymbol{u}}\dot{\boldsymbol{p}}_i^b + \widehat{\boldsymbol{u}}\dot{\boldsymbol{d}}_i + \ddot{\boldsymbol{d}}_i), \quad \boldsymbol{a} = \sum_{i=1}^m \boldsymbol{a}_i, \\
        \boldsymbol{b}_i &= \widehat{\boldsymbol{d}}_i\boldsymbol{a}_i, \quad \boldsymbol{b} = \sum_{i=1}^m \boldsymbol{b}_i.
    \end{align}
\end{subequations}
 With these definitions, $\boldsymbol{f}_{act}$ and $\boldsymbol{l}_{act}$ can be rewritten as
\begin{subequations}
    \begin{align}
        \boldsymbol{f}_{act} &= \boldsymbol{R}(\boldsymbol{a} + \boldsymbol{A}\dot{\boldsymbol{v}} + \boldsymbol{G}\dot{\boldsymbol{u}}), \\
        \boldsymbol{l}_{act} &= \boldsymbol{R}(\boldsymbol{b} + \boldsymbol{B}\dot{\boldsymbol{v}} + \boldsymbol{H}\dot{\boldsymbol{u}}).
    \end{align}
\end{subequations}
Inserting these into \eqref{ForceAndCoupleDecomposition} and using this for the external force and load terms in \eqref{ExplicitModelEquations} yields the complete tendon actuated kinetostatic model with non-uniform stiffness matrices:
\begin{subequations}
\label{ExplicitModelEquationsAugmented}
    \begin{align}
        \dot{\boldsymbol{p}} &= \boldsymbol{R}\boldsymbol{v}, \\
        \label{ExplicitModelEquationsAugmentedR}
        \dot{\boldsymbol{R}} &= \boldsymbol{R}\widehat{\boldsymbol{u}}, \\
        \begin{bmatrix}
            \dot{\boldsymbol{v}}\\
            \dot{\boldsymbol{u}}
        \end{bmatrix} &=
        \begin{bmatrix}
            \boldsymbol{K}_{se} + \boldsymbol{A} & \boldsymbol{G}\\
            \boldsymbol{B} & \boldsymbol{K}_{bt} + \boldsymbol{H}
        \end{bmatrix}^{-1}
        \begin{bmatrix}
            \boldsymbol{d}\\
            \boldsymbol{c}
        \end{bmatrix}.
    \end{align}
\end{subequations}
The auxiliary variables $\boldsymbol{c}$ and $\boldsymbol{d}$ are defined as
\begin{subequations}
\label{ExplicitModelEquationsAugmentedIdentities}
    \begin{align}
        \boldsymbol{c} &= \boldsymbol{K}_{bt}\dot{\boldsymbol{u}}_0 - (\widehat{\boldsymbol{u}}\boldsymbol{K}_{bt} + \dot{\boldsymbol{K}}_{bt})(\boldsymbol{u} - \boldsymbol{u}_0) \nonumber\\
        & \quad - \widehat{\boldsymbol{v}}\boldsymbol{K}_{se}(\boldsymbol{v} - \boldsymbol{v}_0) - \boldsymbol{R}^T\boldsymbol{l}_{ext} - \boldsymbol{b}, \\
        \boldsymbol{d} &= \boldsymbol{K}_{se}\dot{\boldsymbol{v}}_0 - (\widehat{\boldsymbol{u}}\boldsymbol{K}_{se} + \dot{\boldsymbol{K}}_{se})(\boldsymbol{v} - \boldsymbol{v}_0) \nonumber \\ 
        &- \boldsymbol{R}^T\boldsymbol{f}_{ext} - \boldsymbol{a}.
    \end{align}
\end{subequations}
For numerical integration of the boundary value problem, we represent orientation using unit quaternions in \eqref{ExplicitModelEquationsAugmentedR} instead of the rotation matrix $\boldsymbol{R}$. Unit quaternions require only a single normalization condition, making them more numerically stable than rotation matrices, which require more expensive orthogonality corrections at each integration step.


\subsection{Taper Angle Design}
\begin{figure*}[t]
    \vspace{0.2cm}
    \centerline{\includegraphics[width=\linewidth]{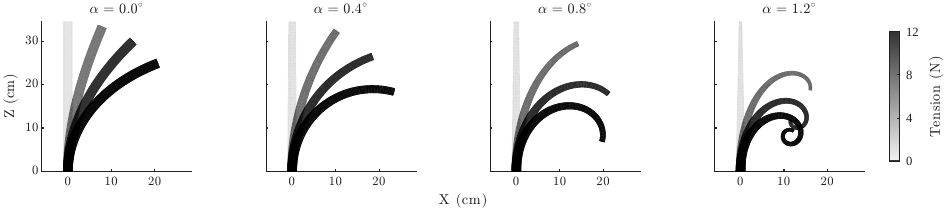}}
    \caption{Simulated backbone shape as a function of taper angle and cable tension. A single cable is tensioned from 0–12 N, modeled using a TPU Young's modulus of 67 MPa, a backbone length of 34.5 cm, and a base radius of 1.11 cm.}
    \label{fig:taper_sweep}
    \vspace{-0.5cm}
\end{figure*}

The kinetostatic model enables systematic exploration of how design parameters, such as taper angle, backbone length, and cable tensions, determine the robot's configuration space, reducing the need for physical prototyping in early design stages. The compliant TPU material and tapered backbone together enable highly curved configurations with large manipulability at modest cable tensions, as demonstrated in \cref{fig:taper_sweep}, where the taper angle $\alpha$ is defined as
\begin{gather}
    \alpha = \arctan\left(\frac{r_\text{base} - r_\text{tip}}{l}\right),
\end{gather}
with $r_\text{base}$ and $r_\text{tip}$ denoting the backbone radii at the base and tip, respectively.

To aid practitioners in finding physical parameters that achieve a desired configuration space, we formulate an inverse design problem that maps a desired curvature profile $\boldsymbol{u}_d(s)$ to the backbone taper angle $\alpha$ using the model from \cref{sec:model_augmentation}. Assuming a circular, linearly tapered backbone, we define the following optimization problem:
\begin{equation}
\label{CostFunction}
    \min_{\alpha}\int_0^l ||\boldsymbol{u}(s,\alpha) - \boldsymbol{u}_d(s)||^2 \ ds.
\end{equation}
The maximum tendon tension the motors are capable of providing is known a priori and is used to constrain the optimization problem by fixing the tendon tension required for obtaining a desired robot curvature. This optimization problem is solved by first choosing an initial guess $\alpha$ and then solving the forward Cosserat boundary value problem. The curvature profile is described by the rotational strain $\boldsymbol{u}(s)$, which is a state variable in \eqref{ExplicitModelEquationsAugmentedIdentities}, and is extracted directly from the model and then inserted into the cost function in \eqref{CostFunction}. A local minimum is then iteratively approached by repeating this procedure.

\begin{figure}[t]
    \centering
    \includegraphics[width=\linewidth]{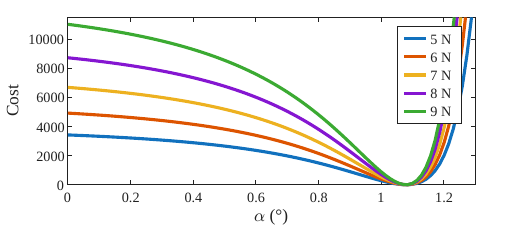}
    \caption{Cost function values over a range of tendon tensions for an optimal taper angle of $\alpha^* = 1.08^\circ$.}
    \label{fig:cost_function}
\end{figure}

\begin{table}[t]
    \vspace{0.15cm}
    \caption{Optimal angle error ($\times 10^{-2}$ degrees) varying tendon tension and taper angle under 50\% element-wise curvature noise.}
    \centering
    \setlength{\tabcolsep}{8pt}
    \begin{tabular}{ccccc}
        \toprule
        & \multicolumn{4}{c}{\textbf{Taper Angle} ($\alpha$)} \\
        \cmidrule(lr){2-5}
        \textbf{Tension (N)} & $0.0^\circ$ & $0.4^\circ$ & $0.8^\circ$ & $1.2^\circ$ \\
        \midrule
        5 & $\phantom{-}0.28$ & $-1.18$ & $\phantom{-}0.24$ & $-0.54$ \\
        6 & $\phantom{-}0.24$ & $-3.47$ & $\phantom{-}1.78$ & $-0.04$ \\
        7 & $\phantom{-}3.41$ & $\phantom{-}0.48$ & $-0.29$ & $-1.68$ \\
        8 & $\phantom{-}0.34$ & $\phantom{-}1.55$ & $-0.59$ & $-3.77$ \\
        9 & $\phantom{-}0.36$ & $-0.19$ & $-1.26$ & $\phantom{-}0.99$ \\
        \bottomrule
    \end{tabular}
    \label{table:inverse_map}
    \vspace{-0.5cm}
\end{table}
\cref{table:inverse_map} shows the errors of the locally optimal taper values $\alpha^*$ over a variation of tendon tensions and taper angles. The other physical parameters are fixed for the optimization, and they are equal to the values used for the experimental model validation in the subsequent section. The desired curvature $\boldsymbol{u}_d(s)$ is found by solving the forward Cosserat boundary value problem with the desired taper angle. To demonstrate robustness to imperfect desired curvature profiles, element-wise random noise of $50\%$ is added to a known nominal curvature profile. The taper angle search space is constrained such that $0^\circ \leq \alpha \leq 2^\circ$. Due to the convex shape of the cost function over the range of candidate taper angles, shown for taper angle $\alpha^* = 1.08^\circ$ in \cref{fig:cost_function}, a local optimization algorithm was chosen. 

\section{MODEL VALIDATION}\label{sec:model_validation}
\begin{figure}[t]
    \centering
    \includegraphics[origin=c, width=\linewidth]{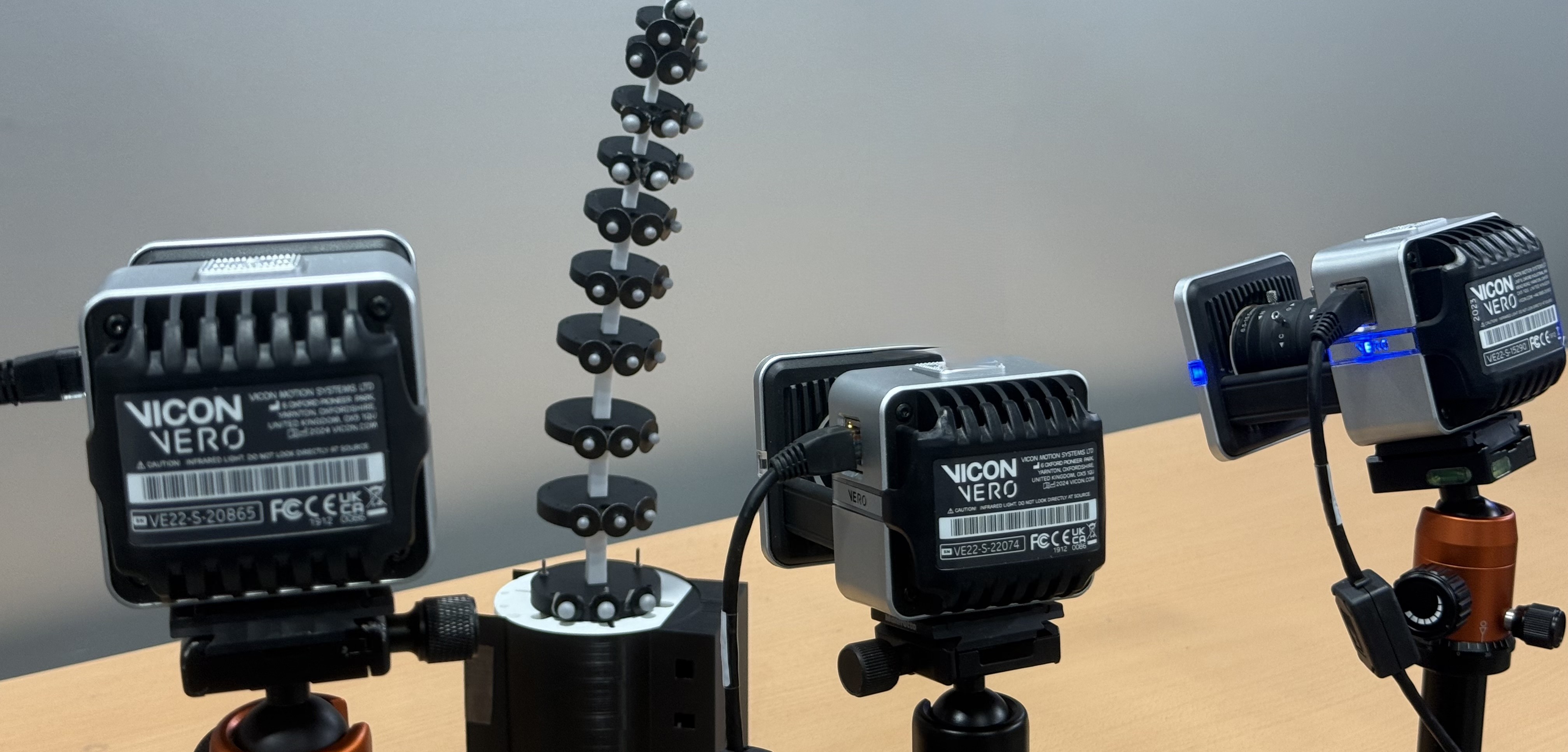}
    \caption{3D-printed tendon-actuated continuum robot with Vicon motion capture markers affixed to the discs' outer circumference.}
    \label{fig:vicon}
    \vspace{-0.5cm}
\end{figure}
In this section, we validate the model presented in \cref{sec:model} against motion capture data from hardware experiments. The robot used for validation was 3D-printed using 1.75 mm TPU 95A filament for the backbone and PLA for the discs, both using a Bambu Lab X1C printer. The backbone had a length of 34.5 cm, a base radius of 1.11 cm, and a tip radius of 0.45 cm, corresponding to a taper angle of approximately $1.08^\circ$. The TPU filament manufacturer reported a Young's modulus of 67 MPa and a Poisson's ratio of 0.39. The backbone is comprised of two 3D-printed segments printed at 100\% infill and bonded together to minimize material parameter uncertainty. While the backbone cross-section was designed to be circular, a squared backbone was introduced during manufacturing to enhance backbone print quality and facilitate proper alignment of the tendon threading holes. To accommodate for this change, the cross-section area ($A(s)$), the second moments of area ($I_{xx}=I_{yy}$), and their derivatives in \eqref{StiffnessMatrices} and \eqref{StiffnessMatricesDot} were modified for squared cross-sections. Moreover, the St. Venant torsion constant of $0.1406$ for a squared solid section was multiplied to the $I_{zz}$ derivation instead of the polar moment.

Ten rigid discs were spaced along the backbone with their radii following a fixed logarithmic ratio and ranging from 3.7 cm at the base to 1.6 cm at the tip. The tendons were routed through the disc holes such that their offset from the backbone center tapers continuously from 3.2 cm at the base to 1.4 cm at the tip. The robot is actuated by three equidistant tendons controlled from the electronics base housing and fastened at the distal disc. The base housing is connected to a nearby computer for power and communication, and each actuator is teleoperated.

The load cells were calibrated using an external force sensor during robot assembly. Each cable was tensioned from 0 to 5 N in 1 N increments, and the corresponding 10-bit ADC readings were recorded to construct a lookup table used for interpolation and extrapolation of tension values. The calibration data per load cell are shown in \cref{table:cali} and exhibit a near-linear relationship, yielding a sensor resolution of approximately 0.125 N/bit. Visual tracking markers were affixed to the outer circumference of the backbone discs, and marker pose measurements were obtained using Vicon Tracker 3.10.2 with Vicon Vero grayscale cameras. The cameras captured the full robot workspace with a position resolution of approximately 0.02 mm. The robot with affixed Vicon markers is shown in \cref{fig:vicon}.
\begin{table}[t]
    \vspace{0.15cm}
    \caption{Load cell calibration: known cable tensions versus 10-bit ADC readings, giving a resolution of approximately 0.125~N/bit over 0--5~N.}
    \centering
    \begin{tabular}{cc cc cc}
        \toprule
        \multicolumn{2}{c}{\textbf{Load Cell 1}} &
        \multicolumn{2}{c}{\textbf{Load Cell 2}} &
        \multicolumn{2}{c}{\textbf{Load Cell 3}} \\
        \cmidrule(lr){1-2} \cmidrule(lr){3-4} \cmidrule(lr){5-6}
        \textbf{Force (N)} & \textbf{Bit} &
        \textbf{Force (N)} & \textbf{Bit} &
        \textbf{Force (N)} & \textbf{Bit} \\
        \midrule
        0.000 & 98  & 0.000 & 100 & 0.000 & 100 \\
        1.079 & 101 & 0.912 & 102 & 0.922 & 104 \\
        1.942 & 110 & 2.099 & 111 & 1.864 & 111 \\
        2.992 & 120 & 3.051 & 118 & 2.884 & 117 \\
        3.953 & 124 & 3.924 & 125 & 4.012 & 125 \\
        5.042 & 142 & 4.993 & 136 & 4.689 & 129 \\
        \bottomrule
    \end{tabular}
    \label{table:cali}
    \vspace{-0.5cm}
\end{table}

The model was validated on a dataset collected by curling and uncurling a single tendon ten times, yielding 1235 sample points with actuated tendon tensions ranging from approximately 2 to 25 N. The original dataset was skewed toward lower tension values, and a bin-based resampling procedure with 1~N bin width was applied to obtain an approximately uniform distribution over tension, preventing lower tensions from dominating calibration. The resampled dataset was randomly split 70/30, with 350 points used for calibration and 150 for testing. To validate the model, load cell measurements were provided as tendon tension inputs to the kinetostatic model, and the predicted backbone positions were compared against Vicon-measured disc positions. However, the two sets of positions are expressed in different coordinate frames, and there exists an unknown constant offset between the markers and backbone, both of which must be accounted for before meaningful comparison is possible.

The optimal transformation between model and Vicon coordinate frames was found using the singular value decomposition method of \cite{Arun1987-od}, which yields the optimal rotation matrix \mbox{$\boldsymbol{R}^* \in \text{SO}(3)$} and translation vector \mbox{$\boldsymbol{t}^* \in \reals{3}$} that minimize $\sum_{i=1}^n \left\| \boldsymbol{R}\, \boldsymbol{r}_{\text{vicon},i}(s) + \boldsymbol{t} - \boldsymbol{r}_{\text{model},i}(s) \right\|^2$, where $n$ is the number of calibration points, $\boldsymbol{r}_{\text{model}, i}(s)$ is the model-predicted backbone position at calibration point $i$, and $\boldsymbol{r}_{\text{vicon}, i}(s)$ is the Vicon marker position at calibration point $i$. Since the Vicon markers are placed on the discs rather than directly on the backbone, they introduce positional offsets that are not precisely known in advance. This unknown offset is approximated by the mean residual after applying the optimal transformation,
\begin{gather}
    \boldsymbol{\delta}(s) = \frac{1}{n} \sum_{i=1}^n \boldsymbol{R}^*\, \boldsymbol{r}_{\text{vicon}, i}(s) + \boldsymbol{t}^* - \boldsymbol{r}_{\text{model}, i}(s).
\end{gather}

Due to nonlinearity and uncertainty in the TPU material properties, the Young's modulus used in the tapered model was set using a tendon tension-scheduling method. The Young's modulus along the backbone varies with strain; however, since strain information is unavailable online, cable tensions are used as a proxy variable to allow for shape-varying Young's modulus values. 
While three tension measurements ($\tau_i \in \reals{}$) are available, the conditioning variables are reduced to the two-dimensional parameter vector \mbox{$\boldsymbol{\Delta} = (\Delta_1, \Delta_2)$}, with $\Delta_1 = \tau_1 - \tau_3$ and $\Delta_2 = \tau_2 - \tau_3$. This parameterization is based on the fact that the differences between cable tensions, rather than their absolute values, determine the shape of the robot.

The tension-scheduled Young's modulus is calibrated on a grid $\mathcal{G} = D_1 \times D_2$, where $D_1$ and $D_2$ are sets of grid nodes spaced at 5~N intervals over the range of tension differences observed during the experiment, with $|D_1| = 5$ and $|D_2| = 3$, giving $|\mathcal{G}| = 15$ calibration points. A modulus value $E^*(\Delta_1, \Delta_2)$ is fitted at each node $(\Delta_1, \Delta_2) \in \mathcal{G}$, and for an arbitrary tension pair the modulus is evaluated by bilinear interpolation over the four surrounding nodes. A global optimization problem is then solved for the entries of $E^*$ using \mbox{MATLAB's} \texttt{surrogateopt} function with default hyperparameters over 200 iterations, with each $E^*$ value constrained between 50 and 200~MPa. At each optimization iteration, the optimal rigid-body transformation was computed, and bias-corrected position residuals were used as the Young's modulus calibration cost function,
\begin{gather} \label{eq:cost}
    \sum_{i=1}^n \left\| \boldsymbol{R}^*\, \boldsymbol{r}_{\text{vicon},i}(s) + \boldsymbol{t}^* - \boldsymbol{r}_{\text{model},i}(s) - \boldsymbol{\delta}(s) \right\|^2.
\end{gather}
This global optimization method was selected over other methods because it is specifically designed for cost functions that are computationally expensive to evaluate.

\begin{figure*}[t]
    \vspace{0.15cm}
    \centering
    \includegraphics[width=\linewidth]{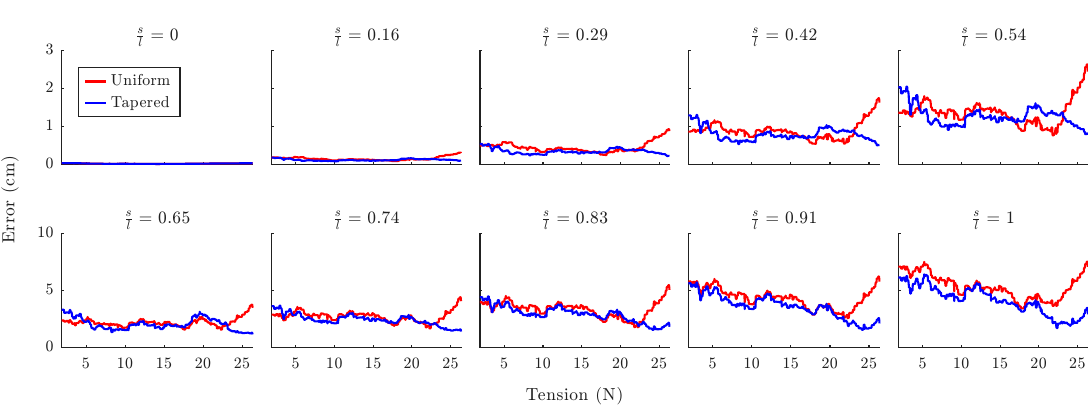}
    \caption{Position error magnitudes of the uniform and tapered models relative to Vicon ground truth, evaluated at multiple points along the backbone. The tapered model uses the calibrated Young's modulus tension-scheduling, while the uniform model uses the manufacturer's value. Optimal transformations and bias corrections for each model were calibrated on the training set. 
    Mean error values were computed over a $\pm 1$~N sliding window on the held-out test set.}
    \label{fig:histograms}
    \vspace{-0.5cm}
\end{figure*}

The resulting position error magnitudes on the held-out test set are shown in \cref{fig:histograms} as a function of normalized backbone length $s/l$ at the ten disc locations, after applying the optimal transformation and bias correction fitted on the training set. The calibrated tapered backbone model is compared to a uniform backbone model with base and tip radius both set to 1.11~cm, using the manufacturer's Young's modulus, and all other physical parameters the same as in the tapered model. The uniform backbone model is also calibrated to determine the optimal transformation between Vicon and model data.

In \cref{fig:histograms}, the uniform and tapered models perform roughly equivalently at lower tensions ($<$ 20~N) across the backbone, suggesting that the robot can be approximated well as a uniform rod in this regime. At higher tensions ($\geq$ 20~N), however, the tapered model shows improved performance, which is the expected gain over the high-strain regions that the uniform model does not capture accurately. Additionally, both models exhibit higher tip error at low cable tensions. The coarse load cell resolution limits detection of subtle shape changes, and mechanical play between the load cells and base housing occasionally introduces a dead zone near the upright, unactuated configuration, where shape changes cannot be resolved without sufficient applied tension. We hypothesize that a primary source of error is the resolution mismatch between the Vicon system (millimeter-level) and the load cells (0.125~N/bit). Efforts are underway to improve sensor integration to mitigate these effects. Additional error sources arising from nonlinear material properties and manufacturing imperfections may be mitigated through further material testing and modeling, or compensated via residual learning methods such as those in \cite{shen2020}.

\section{CONCLUSIONS} \label{sec:conclusion}

This work presented a reproducible, scalable framework for the design, fabrication, modeling, and experimental validation of tendon-actuated continuum robots with tapered TPU backbones. By combining parametric CAD automation, integrated tendon tension sensing, and a generalized Cosserat rod formulation with spatially varying stiffness, the approach bridges rapid soft robot fabrication with physically grounded, model-based analysis. The kinetostatic model accounts for spatially varying cross-sectional geometry, enabling systematic exploration of the configuration space as a function of geometric design parameters and, conversely, inverse mapping from a desired curvature profile to the associated parameters such as the taper angle. This geometry-aware design tool, which moves beyond constant-stiffness assumptions, was used to highlight the mechanical advantages of tapering, including increased distal curvature and improved configuration space coverage at low tendon tensions.

Experimental validation using motion capture and load cell measurements demonstrated consistent centimeter-level agreement between predicted and measured backbone shapes after Young's modulus calibration, confirming that the framework provides reliable shape prediction for compliant, low-cost platforms despite the material variability inherent to 3D-printed TPU. Future work will target higher-resolution tension measurement and encoder–tension fusion for real-time shape estimation and closed-loop control. Overall, the framework establishes an accessible, reproducible pathway from parametric design to controlled tendon actuation, supporting scalable, task-adaptable continuum robots for inspection and manipulation in constrained environments.

\phantomsection
\section*{APPENDIX} \label{appendix}
CAD automation files and Cosserat rod model source code: 
\url{https://github.com/haralmha/CTAM_design_fabrication_modeling}.

Video link: \mbox{\url{https://youtu.be/xW2aTMugGJg}}.

\bibliographystyle{IEEEtran}
\bibliography{CTAM_redesign.bib}{}

\end{document}